\documentclass[conference]{IEEEtran}
\IEEEoverridecommandlockouts
\usepackage[noadjust]{cite}
\usepackage{amsmath,amssymb,amsfonts}
\usepackage{algorithmic}
\usepackage{graphicx}
\usepackage{textcomp}
\usepackage{verbatim}
\usepackage{multirow}
\usepackage{xcolor}
\usepackage{svg}
\usepackage{caption}
\usepackage{subcaption}

\def\BibTeX{{\rm B\kern-.05em{\sc i\kern-.025em b}\kern-.08em
    T\kern-.1667em\lower.7ex\hbox{E}\kern-.125emX}}

\title{\LARGE \bf Prediction of Acoustic Communication Performance for AUVs using Gaussian Process Classification\\
\thanks{This work was supported by the Office of Naval Research via grant N00014-23-1-2345 and the National Oceanic and Atmospheric Administration via award NA22OAR0110191.}% <-this % stops a space
\thanks{Yifei Gao, Harun Yetkin, and Daniel Stilwell are with the Bradley Department of Electrical and Computer Engineering, Virginia Tech, Blacksburg, VA, USA, Harun Yetkin is also with Bartin University. James McMahon is with the Acoustics Division, Code 7130, US Naval Research Laboratory, Washington, D.C. }%
}

\author{
    \IEEEauthorblockN{Yifei Gao, Harun Yetkin, McMahon James, Daniel J. Stilwell}
}
\begin{document}
\maketitle
\thispagestyle{empty}
\pagestyle{empty}

\begin{abstract}
\textbf Cooperating autonomous underwater vehicles (AUVs) often rely on acoustic communication to coordinate their actions effectively. However, the reliability of underwater acoustic communication decreases as the communication range between vehicles increases. Consequently, teams of cooperating AUVs typically make conservative assumptions about the maximum range at which they can communicate reliably. To address this limitation, we propose a novel approach that involves learning a map representing the probability of successful communication based on the locations of the transmitting and receiving vehicles. This probabilistic communication map accounts for factors such as the range between vehicles, environmental noise, and multi-path effects at a given location. In pursuit of this goal, we investigate the application of Gaussian process binary classification to generate the desired communication map. We specialize existing results to this specific binary classification problem and explore methods to incorporate uncertainty in vehicle location into the mapping process. Furthermore, we compare the prediction performance of the probability communication map generated using binary classification with that of a signal-to-noise ratio (SNR) communication map generated using Gaussian process regression. Our approach is experimentally validated using communication and navigation data collected during trials with a pair of Virginia Tech 690 AUVs.

\end{abstract}

\section{Introduction}
Teams of cooperating autonomous underwater vehicles (AUVs) typically use acoustic communication to coordinate their actions.  Underwater acoustic communication is characterized by high bit-error rates due to numerous practical challenges, including multi-path, environmental noise, etc \cite{b14, b10}.  This poses a challenge for teams of AUVs that must plan future communication events in order to coordinate future actions.  One solution  is to assume that communication is deterministic but range-limited, see \cite{b15}, \cite{b16} among many others. When applied to the unreliable underwater acoustic communication channel, however, the assumption that communication events are almost always successful may hold only if the communication range is conservatively small, which limits its use in real-world applications. In contrast, we seek methods of predicting acoustic communication performance that are less conservative and that represent its fundamentally stochastic behavior.  Rather than assuming that communication is always successful so long as the range is conservatively short, we seek methods that predict a high probability of communication success in some locations and at certain ranges, and a lower probability of success at other locations and at longer ranges. 

Predicting acoustic communication performance is a topic that has received limited attention in the literature. The authors in \cite{b10, b17, quattrini2020multi, 8647266, clark2022propeml} employ Gaussian processes to predict received signal strength (RSS) for evaluating communication performance. In our prior work \cite{b12,b13}, we modeled the signal-to-noise ratio (SNR) of successful acoustic communication events using a Gaussian process (GP). However, we note that RSS or SNR do not explicitly address the probability of one vehicle successfully communicating with another vehicle. Indeed, one needs to select a threshold above which the value SNR or RSS corresponds to a successful communication event. On the other hand, in this paper, we seek to explicitly address the probability of successful communication. We intuitively model the probability of successful communication with the corresponding successful and unsuccessful communication events using GP classification, and we evaluate the benefit of this intuition by directly comparing against models generated from SNR using GP regression. To the best of our knowledge, this is the first study that seeks to model acoustic communication performance with a GP classifier.

\begin{figure}[t!]
  \includegraphics[width=3.5in]{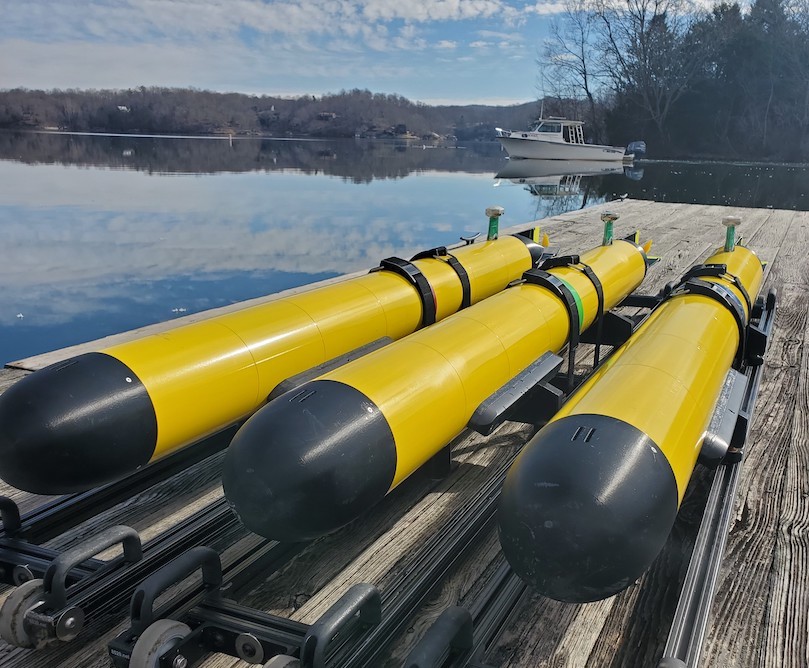}
  \caption{Virginia Tech 690 AUVs}
\label{fig9}
\end{figure}

Our approach is data-driven.  A team of AUVs use occurrences of successful and unsuccessful communication events to compute a map of communication performance. The map is a Gaussian process binary classification of communication success that is used to predict the probability of communication success between an AUV at one location and an AUV at another location.  At the beginning of a mission, the AUVs may assume a simple range-dependent prior.  As data on communication performance is acquired during the mission, the communication performance map evolves, and the AUVs  generate less conservative and more accurate communication performance predictions when planning  their actions. Several Virginia Tech 690 AUVs, which were used for our experiments, are shown in Figure \ref{fig9}

% Hence, in this work, we seek to directly predict the probability of a successful communication event between a transmitting node at one location and a receiving node at another location. Since this approach partitions all events into either being successful or unsuccessful, we propose to model communication performance with GP classification. GP classification is expected to perform better than GP regression since it takes additional unsuccessful communication events into consideration. Indeed, our results demonstrate that accounting for unsuccessful communication events is necessary and that the GP classification methods we propose outperform GP regression in different communication environments, providing motivation for taking unsuccessful communication events into consideration when predicting the communication map.

%In contrast, unsuccessful communication events can be ignored when using Gaussian process regression to model RSS as a spatial function, which is the case in our prior work and that of others, including \cite{b10}, \cite{b17}. 

In work where RSS or SNR is modelled using GP regression, unsuccessful communication events are not explicitly addressed.  Any communication event that results in an SNR or RSS value reported by the acoustic modem can be used for GP regression.  In contrast, modeling communication  success directly using GP classification requires that unsuccessful communication events are explicitly accounted for. The main challenge of accounting for unsuccessful communication events is that the  location of a transmitting node is uncertain.  That is, if a receiving node anticipates the arrival of a data packet from the transmitting node, but it does not arrive, then the receiving node has an uncertain estimate of the location of the transmitting node, which grows with each consecutive unsuccessful communication event. Hence, in this work, we also evaluate the benefit of accounting for input location uncertainty in the communication events when computing a GP classification. We also evaluate the case where location uncertainty is initially unknown but is learned from the data.

Gaussian process classification is used to solve remote sensing image classification for large data streams \cite{8082124} and perform collision checking in the robot motion planning \cite{10137736}, among many other examples. In these specific studies, the location of the data is often assumed to be known. On the contrary, Gaussian process classification with uncertain inputs is rarely addressed in the literature \cite{b1}. There are several methods that address the case of classification with uncertain inputs such as those proposed by \cite{saez2014analyzing}, \cite{feng2018reinforcement} and \cite{7159100}. However, these approaches do not employ Gaussian process classification since we believe one of the primary challenges with Gaussian process classification is the analytical inconvenience on its logit likelihood function. Our proposed sparse classification methods are built upon the previous work on multi-class Gaussian process classification with noisy inputs (NIMGPC) in \cite{b1}, which accounts for multi-class classification using Gaussian processes along with noisy uncertain transmitter locations, and upon \cite{b18} which addresses sparse Gaussian process classification in the variational distribution setting. NIMGPC admits a sparse Gaussian process, which is used for large amounts of data.  In our case, a sparse GP is useful because of the very low bandwidth  acoustic communication channel available underwater. Sharing a small number of inducing points for the sparse GP may be possible, while sharing data of the full GP would be prohibitively difficult in real-time \cite{b11}.  
%Although sharing sparse GP representations among AUVs is not addressed herein, it is a necessary step in future work. To simplify the presentation of numerical examples, we assume that all AUVs are in the same horizontal plane and ignore vertical displacement.  %We use 2D communication locations between AUVs as our training data but actually real locations among AUVs are never known \cite{b9}. Therefore, we take advantage of noisy input framework proposed on NIMGPC \cite{b1} and modify the framework to adapt to binary classification case. 

\noindent The main contributions of this work are
\begin{itemize}
    \item We provide a novel data-driven approach to predict the probability of communication of success suitable for underwater acoustic communication even in the case where the location of the transmitting node is uncertain, as in the case of AUVs. Moreover, our approach explicitly accounts for unsuccessful communication events and the uncertainty in transmitter location that arises when a communication event is unsuccessful.
    \item We show empirically that GP classification outperforms GP regression on communication data acquired using the Virginia Tech 690 AUVs.
    \item We show that accounting for the uncertainty in the location of the transmitter vehicle further improves the results for AUV communication performance over general sparse Gaussian process classification.
\end{itemize}
This work does not seek to specifically address decentralized GP classification for modelling acoustic communication performance.  Rather, we seek to authoritatively evaluate  GP classification performance for teams of underwater robots.  Follow-on work will explicitly address the additional challenges of decentralized classification among a team of underwater robots.

The remainder of this paper is as follows. In Section II, we give a brief review of Gaussian process, sparse Gaussian processes and Gaussian process binary classification. In Section III, we address the problem formulation for noisy input sparse Gaussian process binary classification and derive the evidence lower bound of neural network method modified from the NIMGPC framework to address transmitter location uncertainty. In Section IV, we use communication data acquired from experiments with Virginia Tech 690 AUVs to compare prediction performance between Gaussian process regression methods based on SNR measurements and proposed Gaussian process classification methods.

\section{Preliminaries}
In this section, we briefly review mathematical background of Gaussian process, binary classification and sparse Gaussian process. For additional details, please refer to \cite{b3, b2, b18, hensman2014scalable}.

\subsection{Gaussian Process Regression}
In the standard Gaussian process regression, we consider learning a function $f : \mathbb{R}^{d} \rightarrow \mathbb{R}$ from measurement pairs $\{\textbf{X}, \textbf{y}\}$ where $\textbf{X} = [\textbf{x}_{1},..., \textbf{x}_{N}]^{T}$ and $\textbf{y} = [y_{1},..., y_{N}]^{T}$. We have each measurement $y = f(\textbf{x}) + \epsilon$ which $\textbf{x} \in \mathcal{X} \subset \mathbb{R}^{d}$ and $\epsilon \sim \mathcal{N}(0, \sigma^{2})$ is the measurement noise. We assume that the latent function $f$ is a Gaussian process denoted by $GP(\mu_{N}(\textbf{x}), k_{N}(\textbf{x}, \textbf{x}'))$ where $\bar{f}$ denotes the mean function and $k: \mathcal{X} \times \mathcal{X} \rightarrow \mathbb{R}$ is a kernel function. Given a prior mean and kernel functions $GP(\mu_{0}(.), k_{0}(.))$ and acquired data $\{\textbf{X}, \textbf{y}\}$, GP regression generates a mean and covariance function
\begin{equation}
    \begin{split}
        &\mu_{N}(\textbf{x}) = \textbf{k}(\textbf{x})^{T}(\textbf{K} + \sigma^{2}\textbf{I})^{-1}\textbf{y} \\
        &k_{N}(\textbf{x},\textbf{x}') = k_{0}(\textbf{x},\textbf{x}') - \textbf{k}(\textbf{x})^{T}(\textbf{K} + \sigma^{2} \textbf{I})^{-1} \textbf{k}(\textbf{x}')
    \end{split}
\end{equation}\label{gp1}
where $\textbf{k}(\textbf{x}) \in \mathbb{R}^{N}$, $\textbf{K} \in \mathbb{R}^{N \times N}$, $[\textbf{k}(\textbf{x})]_{i} = k_{0}(\textbf{x}, \textbf{x}_{i})$ for any $\textbf{x}_{i} \in \textbf{X}$, $[\textbf{K}]_{ij} = k_{0}(\textbf{x}_{i}, \textbf{x}_{j})$ and $\textbf{x}, \textbf{x}' \in \mathcal{X}$.

\subsection{Sparse Gaussian Process}
The cost of computing the posterior distribution $p(f_{*} | \textbf{y}, \textbf{x}_{*})$ in \eqref{eq:posterior_fstar} is $\mathcal{O}(N^{3})$ which results from inverting the kernel matrix $\textbf{K}$. Sparse Gaussian process proposed in~\cite{b3} reduces the computational cost to $\mathcal{O}(M^{3})$ where $M \ll N$ is the number of inducing points. We denote the inducing points as $\textbf{Z} = \left[ \textbf{z}_{1},...,\textbf{z}_{M}\right]^{T}$ where each $\textbf{z}_{i}$ has same input dimension $d$. Accordingly, the inducing output vector is $\textbf{f}_{s} = \left[f_{s}(\textbf{z}_{1}),...,f_{s}(\textbf{z}_{M})\right]^{T}$, and
\begin{equation}
    p(\textbf{f} | \textbf{f}_{s}) = \mathcal{N}(\textbf{f} | \textbf{K}_{\textbf{X}, \textbf{Z}}\textbf{K}_{\textbf{Z}, \textbf{Z}}^{-1}\textbf{f}_{s}, \textbf{K}_{\textbf{X}, \textbf{X}} - \textbf{K}_{\textbf{X}, \textbf{Z}}\textbf{K}_{\textbf{Z}, \textbf{Z}}^{-1}\textbf{K}_{\textbf{X}, \textbf{Z}}^{T}) \label{eq3}
\end{equation}
where $\textbf{K}_{\textbf{X},\textbf{Z}}$ is an $N \times M$ covariance matrix and $\textbf{K}_{\textbf{Z},\textbf{Z}}$ is an $M \times M$ covariance matrix. In the variational distribution setting, we propose variational distribution $q(\textbf{f}_{s}) = \mathcal{N}(\boldsymbol{\mu}_{s}, \boldsymbol{\Sigma}_{s})$. Parameters $\boldsymbol{\mu}_{s}$ and $\boldsymbol{\Sigma}_{s}$ are found during the learning process which is addressed in Section III. We assume joint variational distribution $q(\textbf{f},\textbf{f}_{s}) = p(\textbf{f} | \textbf{f}_{s})q(\textbf{f}_{s})$. Then, the distribution $p(\textbf{f} | \textbf{y})$ is approximated as in \cite{b1} and \cite{b3},
\begin{equation}
    p(\textbf{f} | \textbf{y}) \approx \int p(\textbf{f} | \textbf{f}_{s})q(\textbf{f}_{s}) d\textbf{f}_{s} = \mathcal{N}(\boldsymbol{\mu}, \boldsymbol{\Sigma}) \label{eq4}
\end{equation}
where
\begin{align}
    \boldsymbol{\mu} &= \textbf{K}_{\textbf{X}, \textbf{Z}}\textbf{K}_{\textbf{Z}, \textbf{Z}}^{-1}\boldsymbol{\mu}_{s} \label{eq5}\\
    \boldsymbol{\Sigma} &= \textbf{K}_{\textbf{X},\textbf{X}} - \textbf{K}_{\textbf{X}, \textbf{Z}}\textbf{K}_{\textbf{Z},\textbf{Z}}^{-1}\textbf{K}_{\textbf{X},\textbf{Z}}^{T} + \textbf{K}_{\textbf{X},\textbf{Z}}\textbf{K}_{\textbf{Z},\textbf{Z}}^{-1}\boldsymbol{\Sigma}_{s}\textbf{K}_{\textbf{Z},\textbf{Z}}^{-1}\textbf{K}_{\textbf{X},\textbf{Z}}^{T} \label{eq6}
\end{align}

\subsection{Gaussian Process Binary Classification}
In the Gaussian process binary classification case, we have the corresponding observed target labels for each instance $\textbf{y} = \left[y_{1},...,y_{N}\right]^{T}$ where $y_{i} \in \{0,1\}$. Since it is a binary classification problem, it suffices to compute
\begin{equation}%\label{eq1}
p(y_{*} = 1| \textbf{y}, \textbf{x}_{*}) = \int p(y_{*} = 1| f_{*})p(f_{*} | \textbf{y}, \textbf{x}_{*}) df_{*}
\label{eq:posterior_fstar}
\end{equation}
\noindent where $\textbf{x}_{*}$ is a test input, $y_* = 1$ denotes predicting a successful communication event, $f_{*}$ is the latent function that we are trying to learn such that $f(\textbf{X}) \sim \mathcal{GP}(\textbf{0}, \textbf{K})$ where $[\textbf{K}]_{ij} = k_{0}(\textbf{x}_{i},\textbf{x}_{j})$, and $p(y_{i} | f_{i})$ is the Bernoulli distribution
\begin{equation}
p(y_{i} | f_{i}) = \left(\frac{e^{f_{i}}}{1 + e^{f_{i}}}\right)^{y_{i}}\left( 1 - \frac{e^{f_{i}}}{1 + e^{f_{i}}}\right)^{(1 - y_{i})} \label{eq2}
\end{equation}

The posterior distribution $p(f_{*} | \textbf{y}, \textbf{x}_{*})$ in~\eqref{eq:posterior_fstar} is non-Gaussian \eqref{gp1} but can be approximated as a Gaussian distribution using the existing approximation inference methods such as variational inference \cite{b3} or expectation propagation \cite{b4}.

%%%%%%%%%%%%%%%%%%%%%%%%%%%%%%%%%%%%%%%%%%%%
\section{Binary Classification with Noisy Inputs}
\label{sec:methods}
In this section, we describe the problem formulation that accounts for noisy inputs due to the uncertainty in the location of the transmitter vehicle. We build upon the neural network in \cite{b1} and we summarize the necessary results. For more details, we refer the reader to \cite{b1}. Here we provide the derivation of the evidence lower bound for the neural network method being upper bounded by the log likelihood, and we derive an expression for the  predicted classification probability at a test input (location) value.
%We also give evidence lower bound of first-order method explicitly which is not shown in \cite{b1} and make it adapt to binary case.

We consider a set of known noisy inputs $\tilde{\textbf{X}} = \left[\tilde{\textbf{x}}_{1},..., \tilde{\textbf{x}}_{N}\right]^{T}$ and a corresponding set of unknown true inputs $\textbf{X}$. That is, 
\begin{equation}
    \tilde{\textbf{x}}_{i} = \textbf{x}_{i} + \boldsymbol{\epsilon}_{i} \quad \text{such that} \quad \boldsymbol{\epsilon}_{i} \in \mathcal{N}(\textbf{0}, \boldsymbol{V}_{i})
    \label{eq:noisyinput}
\end{equation}
where $\textbf{x}_{i}$ is the true input, and $\boldsymbol{V}_{i}$ is a $d \times d$ diagonal covariance matrix \cite{b1, b6}. In Subsection \ref{subsec.elbnn} we describe how the covariance matrix $\boldsymbol{V}_{i}$ is learned from the training data.

\subsection{Evidence Lower Bound of Neural Network Method} \label{subsec.elbnn}

In GP classification, the evidence lower bound (ELBO)  serves as a function that can be maximized to find the optimized kernel hyper-parameters just like maximizing the marginal likelihood in GP regression \cite{b2}. Similarly, in sparse GP classification, we can also use the evidence lower bound to find inducing points locations. The evidence lower bound is originally addressed in \cite{jordan1999introduction}. Here we derive the evidence lower bound of the neural network method for the binary classification case to learn all parameters of interest including noise covariance matrix $\boldsymbol{V}_{i}$, kernel hyper-parameters, inducing points $\textbf{Z}$ , variational distribution $q(\textbf{f}_{s}) = \mathcal{N}(\boldsymbol{\mu}_{s}, \boldsymbol{\Sigma}_{s})$ and neural network parameters $\boldsymbol{\theta}$ which are indicated below. In general, we seek parameters of the approximate posterior $q(\textbf{X},\textbf{f},\textbf{f}_{s})$ such that it minimizes approximation error with respect to KL divergence. That is,
\begin{align}
\underset{\{l_{i}\}_{i = 1}^{d},\sigma_{f}^{2},\sigma_{0}^{2},\boldsymbol{\mu_{s}},\boldsymbol{\Sigma}_{s},\textbf{Z}, \boldsymbol{\theta} , \textbf{V}_{i}}{\arg\min} KL(q(\textbf{X},\textbf{f},\textbf{f}_{s}) | p(\textbf{X},\textbf{f},\textbf{f}_{s} | \textbf{y}, \tilde{\textbf{X}})) \label{eq12-1}
\end{align}
where $\{\{l_{i}\}_{i = 1}^{d},\sigma_{f}^{2},\sigma_{0}^{2}\}$ are radial basis kernel hyper-parameters shown in Section IV. 
 Similar to [3], neural network parameters $\boldsymbol{\theta}$ parameterize mean and covariance matrix of variational distribution $q(\textbf{x}_{i})$ 
\begin{align}
    q(\textbf{x}_{i}) = \mathcal{N}(\textbf{x}_{i} | \boldsymbol{\mu}_{\theta}(\tilde{\textbf{x}}_{i}, y_{i}), \textbf{V}_{\theta}(\tilde{\textbf{x}}_{i},y_{i}))
\end{align} \label{eq12-2}
where $y_{i} \in \{0,1\}$. And $p(\textbf{X},\textbf{f},\textbf{f}_{s} | \textbf{y}, \tilde{\textbf{X}})$ and $q(\textbf{X},\textbf{f},\textbf{f}_{s})$ are the exact posterior and approximated posterior which in this case are same as in  \cite{b1}. However, directly minimizing the KL divergence objective function is intractable since the posterior distribution is intractable. Instead, we show that the parameters can be equivalently optimized by maximizing the evidence lower bound. By expanding the KL divergence objective using posterior $p(\textbf{X},\textbf{f},\textbf{f}_{s} | \textbf{y}, \tilde{\textbf{X}})$ expression in \cite{b1} and \eqref{eq12-1} based on the definition of KL divergence, we can show that
\begin{multline}
        KL(q | p) = E_{q} \,[ \log q(\textbf{X},\textbf{f},\textbf{f}_{s})\,] - E_{q} \,[ \log p(\textbf{X},\tilde{\textbf{X}},\textbf{y},\textbf{f},\textbf{f}_{s})\,] \\
        + \log p (\textbf{y}, \tilde{\textbf{X}}) \label{eq13}
\end{multline}
Here we use the notational shorthand $q = q(\textbf{X}, \textbf{f}, \textbf{f}_{s})$. We define neural network evidence lower bound as $ELBO_{NN} = E_{q} \left[\log\frac{p(\textbf{X},\tilde{\textbf{X}},\textbf{y},\textbf{f},\textbf{f}_{s})}{q(\textbf{X},\textbf{f},\textbf{f}_{s})}\right]$ \cite{b1}. After expanding the term, we have
\begin{multline}
    ELBO_{NN} = \sum_{i = 1}^{N}E_{q} \,[ \log p(y_{i} | f_{i})\,] + \sum_{i = 1}^{N} E_{q} \,[ \log p(\tilde{\textbf{x}}_{i} | \textbf{x}_{i})\,]\\ 
    - KL(q(\textbf{f}_{s} | p(\textbf{f}_{s}))) - \sum_{i = 1}^{N} KL(q(\textbf{x}_{i} | p(\textbf{x}_{i}))) 
    \label{eq:elbo_nn}
\end{multline}
where the term $E_{q}[\log p(y_{i} | f_{i})]$ in~\eqref{eq:elbo_nn}, can be approximated as
\begin{multline}
   E_{q} \,[ \log p(y_{i} | f_{i})\,]
    \approx \frac{1}{M}\sum_{i = 1}^{M}\sum_{j = 1}^{m} w_{j} \frac{1}{\sqrt{\pi}} \\
    \log \left(\left(\frac{e^{\sqrt{2} \Sigma_{ii} t_{j} + \mu_{i}}}{1 + e^{\sqrt{2} \Sigma_{ii} t_{j} + \mu_{i}}}\right)^{y_{i}}
    \left( 1 - \frac{e^{\sqrt{2} \Sigma_{ii} t_{j} + \mu_{i}}}{1 + e^{\sqrt{2} \Sigma_{ii} t_{j} + \mu_{i}}}\right)^{(1 - y_{i})}\right) \label{eq32}
\end{multline}
where $t_{j}$, $w_{j}$ are sample locations and sample weights respectively, and $m$ is the number of grid elements for Gaussian quadrature. We have $\mu_i$ and $\Sigma_{ii}$ 
\begin{align}
    \mu_{i} &= \textbf{K}_{\textbf{x}_{i},\textbf{Z}}\textbf{K}_{\textbf{Z},\textbf{Z}}^{-1}\boldsymbol{\mu}_{s} \label{eq33}
\end{align}
and  
\begin{align}
    \Sigma_{ii} = \sqrt{\textbf{K}_{\textbf{x}_{i},\textbf{x}_{i}} - \textbf{K}_{\textbf{x}_{i},\textbf{Z}}\textbf{K}_{\textbf{Z},\textbf{Z}}^{-1}\textbf{K}_{\textbf{x}_{i},\textbf{Z}}^{T} + \textbf{K}_{\textbf{x}_{i},\textbf{Z}} \textbf{K}_{\textbf{Z},\textbf{Z}}^{-1}\boldsymbol{\Sigma}_{s} \textbf{K}_{\textbf{Z},\textbf{Z}}^{-1}\textbf{K}_{\textbf{x}_{i},\textbf{Z}}^{T}}\label{eq34}
\end{align}
Each sample $\textbf{x}_{i}$ is generated from the variational distribution $q(\textbf{x}_{i})$ using Monte Carlo combined with the reparametrization trick, and $y_{i} \in \{0, 1\}$ is the corresponding training label for $\textbf{x}_{i}$. We note that $ELBO_{NN} \leq \log p(\textbf{y}, \tilde{\textbf{X}}) $ since $KL(q | p)$ is always positive. We learn the neural network parameters $\boldsymbol{\theta}$ based on maximizing $ELBO_{NN}$.

\subsection{Prediction}
After optimizing the $ELBO$, we can predict the probability $p(y_{*} | \tilde{\textbf{x}}_{*}, \textbf{y})$ which is the approximate predictive distribution for class label $y_{*}$ of a new instance $\tilde{\textbf{x}}_{*}$ given all training labels $\textbf{y}$,
\begin{equation}
    p(y_{*} | \tilde{\textbf{x}}_{*}, \textbf{y}) \approx \int p(y_{*} | f_{*}) \left[p(f_{*} |\textbf{f}_{s}) q(\textbf{f}_{s}) d\textbf{f}_{s} \right] p(\textbf{x}_{*} | \tilde{\textbf{x}}_{*}) d\textbf{x}_{*}df_{*}\label{eq23}
\end{equation}
where $p(\textbf{x}_{*} | \tilde{\textbf{x}}_{*})$ is the posterior distribution of the actual attributes of the new instance given the observed attributed $\tilde{\textbf{x}}_{*}$, which is shown in \cite{b1}.
For approximating the integral \eqref{eq23}, we can generate samples of $\textbf{x}_{*}$ simply by drawing from $p(\textbf{x}_{*} | \tilde{\textbf{x}}_{*})$  to compute a Monte Carlo approximation,
\begin{align}
    p(y_{*} | \tilde{\textbf{x}}_{*}, \textbf{y}) \approx \frac{1}{M} \sum_{m = 1}^{M}\int p(y_{*} | f_{*}^{m}) \left[p(f_{*}^{m} |\textbf{f}_{s}) q(\textbf{f}_{s}) d\textbf{f}_{s} \right] df_{*}^{m} \label{eq25}
\end{align}
Since the closed-form solution of the integral is a Gaussian distribution,
\begin{equation}
    \int p(f_{*}^{m} |\textbf{f}_{s}) q(\textbf{f}_{s}) d\textbf{f}_{s} = \mathcal{N}(\mu_{*}^{m}, \Sigma_{*}^{m}) \label{eq26}
\end{equation}
where 
\begin{align}
    \mu_{*}^{m} &= \textbf{K}_{\textbf{x}_{*}^{m},\textbf{Z}}\textbf{K}_{\textbf{Z},\textbf{Z}}^{-1}\boldsymbol{\mu}_{s},  \label{eq27}
\end{align}
\begin{multline}
    \Sigma_{*}^{m} = \\
    \textbf{K}_{\textbf{x}_{*}^{m}, \textbf{x}_{*}^{m}} - \textbf{K}_{\textbf{x}_{*}^{m}, \textbf{Z}}\textbf{K}_{\textbf{Z},\textbf{Z}}^{-1}\textbf{K}_{\textbf{x}_{*}^{m}, \textbf{Z}}^{T} + \textbf{K}_{\textbf{x}_{*}^{m},\textbf{Z}}\textbf{K}_{\textbf{Z},\textbf{Z}}^{-1}\boldsymbol{\Sigma}_{s}\textbf{K}_{\textbf{Z},\textbf{Z}}^{-1}\textbf{K}_{\textbf{x}_{*}^{m},\textbf{Z}}^{T},  \label{eq28}
\end{multline}
and $\textbf{x}_{*}^{m}$ is one generated test sample from the distribution $p(\textbf{x}_{*} | \tilde{\textbf{x}}_{*})$. We note that all the parameters are already optimized based on maximizing $ELBO$. Therefore, 
\begin{align}
    p(y_{*} | \tilde{\textbf{x}}_{*}, \textbf{y}) &\approx \frac{1}{M}\sum_{m = 1}^{M}\int p(y_{*}| f_{*}^{m}) \mathcal{N}(\mu_{*}^{m}, \Sigma_{*}^{m}) df_{*}^{m} \label{eq29}
\end{align}
We note that this integral can be approximated by Gaussian one-dimensional quadrature.

\section{Numerical Results}

We evaluate our approach to predict
acoustic communication performance using three real datasets
acquired from Virginia Tech 690 AUVs in Claytor Lake, near
Dublin, Virginia, USA. The 690 AUV displaces approximately 43Kg and is 2.23 meters long.  It communicates acoustically using the WHOI micromdem 2, which operates at 25KHz.  For these experiments, the packet size was 36 bytes. The estimates its location uses an inertial navigation system aided by a Doppler velocity log while underwater, and GPS while at the surface.  The inertial navigation algorithm estimates localization uncertainty.

Each AUV records the time and location of received and transmitted communication events. Furthermore, each AUV transmits on a predetermined schedule so that the other vehicle can infer when an unsuccessful communication event has occurred. To reduce complexity we assume that both AUVs are in the same horizontal plane and ignore vertical displacement. The analysis contains two parts. In the first part, we compare our proposed classification methods with regression methods on different environments and demonstrate that classification methods outperform regression on all communication environments averaged over 20 validation data sets. 
For each dataset (e.g., \textbf{DataSet1}), these 20 distinct validation sets are randomly generated. Each validation set comprises 20 percent of the entire dataset and includes a balanced mix of successful and unsuccessful communications. This approach ensures an unbiased representation of the data.

The remaining 80 percent of the dataset, which includes both successful and unsuccessful events, is used to form the training set for Gaussian process classification. This split allows for a comprehensive evaluation of the classifier's performance. Furthermore, within each classification training set, we identify and extract the data points corresponding to successful communications. These selected points and their corresponding SNR values serve as the training set for Gaussian process regression. We also show that accounting for input noise can lead to both an increase in communication ratio and a decrease negative log-likelihood (NLL) on Gaussian process classification, which are the two metrics we use to evaluate performance In the second part, we extract one validation set from 20 validation sets and compare communication prediction performance between classification and regression on the environment.

We provide three typical data sets for performance evaluation. \textbf{DataSet1} contains 72 successful and 144 unsuccessful communication events, indicating a poor communication quality environment. \textbf{DataSet2} comprises 120 successful and 135 unsuccessful communication events. The final \textbf{DataSet3} contains 311 successful communication points and 103 unsuccessful communication points, signifying a more favorable communication quality environment.

For training inputs in both regression and classification methods, we extract the corresponding communication locations between two autonomous underwater vehicles (AUVs) in 2D, so each input $\textbf{x} \in \mathbb{R}^4$. Outputs for all Gaussian process (GP) regression methods are the predicted mean signal-to-noise ratio (SNR) values in decibels (dB) and variances at the corresponding test location $\textbf{x}_{*}$. Outputs for all GP classification methods are the predicted probability of the test label being 1, which is $p(y_{*} = 1 | \textbf{x}_{*})$ given the test location $\textbf{x}_{*}$.

We begin by comparing the communication prediction performance between Gaussian process regression and classification. For both GP regression on SNR and GP classification on probability, a threshold  is required above which a communication event is expected to be successful. We identify the optimal signal-to-noise ratio threshold of $9.8dB$, which achieves the highest average communication ratio in mission 1 across 20 different validation sets. The communication performance is then evaluated using this threshold on the remaining two missions. Furthermore, we compare the $9.8dB$ threshold, derived from the first data set, with the best thresholds computed on the other two missions. For the case of GP classification, a probability of communication success greater than 0.5 is predicted to be a successful communication event. It is important to note that for regression, only the predicted mean is compared against the specific threshold, disregarding the expected variance. Similarly, while classification methods also compute latent variance, we do not consider variance in either regression or classification.

On all the tables shown, we define $ratio$ as the following,
\begin{equation}
    ratio_{threshold} = \frac{n_{TP} + n_{FN}}{N_{total}}
\end{equation}
where $n_{TP}$ is the number of corrected communicating attempts if GP prediction is higher than the defined threshold and $n_{FN}$ is the number of corrected not communicating attempts if GP prediction is lower than the defined threshold. And $N_{total}$ is the total number of validation points. For classification metric, we also use negative log likelihood
which is defined
\begin{equation}
    NLL = - \frac{1}{N_{total}}\sum_{i = 1}^{N_{total}} y_{i} \log(p_{i}) + (1 - y_{i}) \log(1 - p_{i})
\end{equation}
where $y_{i}$ is the $i$th true label in the validation data set. And $p_{i}$ gives probability prediction on the specific label $y_{i}$. Hence, metric $NLL$ quantifies certainty about that data points in the classification compared with the $ratio$ metric.

For all methods provided, we model the Gaussian process using the squared exponential kernel with automatic relevance determination \cite{b6, b1},
\begin{align}
    k(\textbf{x}, \textbf{x}') = \sigma_{f}^{2} \exp\left[- \frac{1}{2} \sum_{i = 1}^{d} \frac{(x_{i} - x_{i}')}{l_{i}}\right] + \sigma_{0}^{2} I[\textbf{x} = \textbf{x}'] \label{eq30}
\end{align}
where $\sigma_{f}^{2}$ and $\sigma_{o}^{2}$ are the signal variance and the additive Gaussian noise variance, respectively, and $\{l_{i}\}_{i = 1}^{d}$ are length-scales accounting for each input dimension. The indicator function $I[\cdot]$ evaluates to 1 if and only if the two inputs $\textbf{x}$ and $\textbf{x}'$ are equal.

Similar to the method adopted in \cite{b1}, we uniformly select a constant mini-batch size of training samples from each whole \textbf{DataSet} with a size of 50 without replacement, where each training sample contains the 2-D positions
for both AUVs. For the neural network method, we maximize the ELBO with epoch =
1000 and one hidden layer with fifty hidden units are used for all experiments. We choose the number of inducing points to be five percent of the input training data in each sparse method and initialize these inducing points on training points.

We employ GPy \cite{gpy2014} to compute the first three columns in the Table \ref{tab.Mission1}, Table \ref{tab.Mission2}, and Table \ref{tab.Mission3}, corresponding to standard Gaussian process regression (GPR), sparse variational Gaussian process regression (SVGPR), and Gaussian process classification (GPC). For each mission and method, we utilize the same random seed for 20 validation sets to facilitate consistent performance comparison across all provided methods. The size of each validation set constitutes 20 percent of the entire data set for the respective mission. Consequently, each table value presented for a given mission represents the average over the same 20 validation sets. For the last two columns, we employ a modified version of binary classification from \cite{b1}, which includes sparse variational Gaussian process binary classification (SVGPC) and noise-input neural network (NI-NN) for binary classification.We conduct training for SVGPC and NI-NN methods using an Nvidia T4 GPU. The average training time varies across three datasets. For DataSet1, SVGPC requires 222.2 seconds, while NI-NN takes 263.6 seconds. On DataSet2, the average training time increases to 270.5 seconds for SVGPC and 320.5 seconds for NI-NN. DataSet3 demands the longest training periods with a relatively large number of data points, with SVGPC averaging 361.8 seconds and NI-NN requiring 439.0 seconds.

Before evaluating each validation set, we give the same successful training points for both regression and classification methods. However, classification methods can acquire additional unsuccessful training points. Hence, in each dataset for each validation set, classification methods have the same successful training points as regression methods have, plus additional unsuccessful training points depending on the size of the whole data. For the first and second missions ( Table \ref{tab.Mission1} and table \ref{tab.Mission2} ), all the Gaussian process classification methods outperform two regression methods computed from Gpy using the threshold $9.8dB$ and $10.2dB$, which are the best thresholds on missions 1 and 2, respectively. \textbf{In fact, we find GPC is at least as competitive as GPR and SVGPR on the ratio metric for all missions no matter what threshold we set for regression methods on different communication data sets we provide}. We also demonstrate on all missions in different environments that taking noisy inputs into consideration has the potential to improve both communication ratio accuracy and decrease negative log-likelihood, which is indicated in the last column for each mission for the neural network method we propose. Note that in noisy input case we only compare the derived neural network method with sparse variational Gaussian process classification(SVGPC) just like in \cite{b1}. For \textbf{DataSet2}, $10.2dB$ is the best threshold we find for the second mission, but it is still not competitive with any classification method. In \textbf{DataSet2}, sparse methods actually outperform GPC but have a trade-off in increasing the negative-log likelihood.

There is an interesting observation in \textbf{DataSet3}: as the data becomes increasingly dominated by successful communication events, we can always choose a lower threshold for the regression methods GPR and SVGPR to achieve a high $ratio$ since regression only accounts for successful communication events ( Table \ref{tab.Mission3} ). And it turns out that $0.1dB$ is the best threshold if we use the $ratio$ metric. Hence, we choose $0.1dB$, which is also the lowest signal-to-noise ratio (SNR) measurement in the data set for \textbf{DataSet3}, and we find that the GPC method is still at least competitive with GPR. We demonstrate that if we set the threshold to $9.8dB$, which is the best threshold found for mission 1, we achieve a relatively low $ratio$ for GPR and SVGPR. This is quite significant for future real-time analysis, as it shows that it is challenging to find a consistently good threshold for GPR.  In contract, for GP classification, the result explicitly represents probability of successful communication, so no threshold is needed.

\begin{figure}[t!]
\centering
  \begin{subfigure}[b]{0.35\textwidth}
  \centering
  \includegraphics[width=\textwidth]{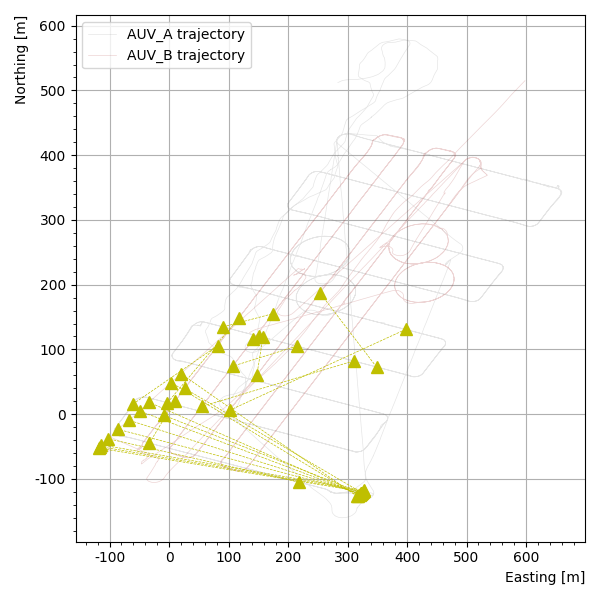}
  \end{subfigure}
  \begin{subfigure}[b]{0.35\textwidth}
  \centering
  \includegraphics[width=\textwidth]{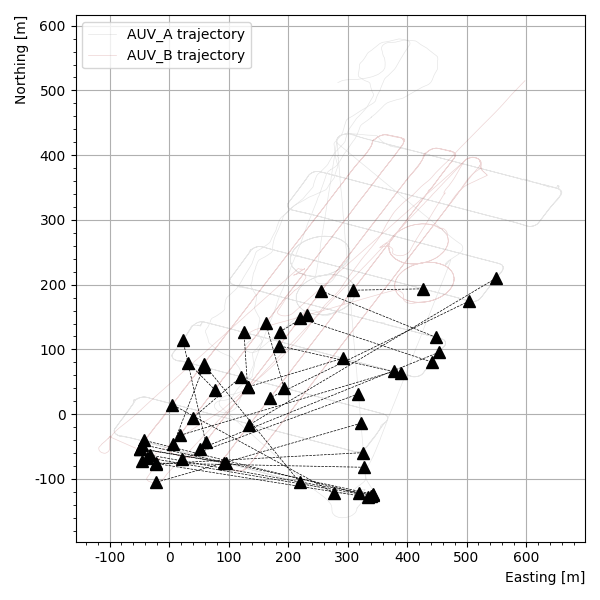}
  \end{subfigure}  
\caption{Successful (top) and unsuccessful(bottom) ground truth validation points}
\label{fig5}
\end{figure}

\begin{figure}[t!]
    \centering
  \begin{subfigure}[b]{0.23\textwidth}
  %\centering
  \includegraphics[width=\textwidth]{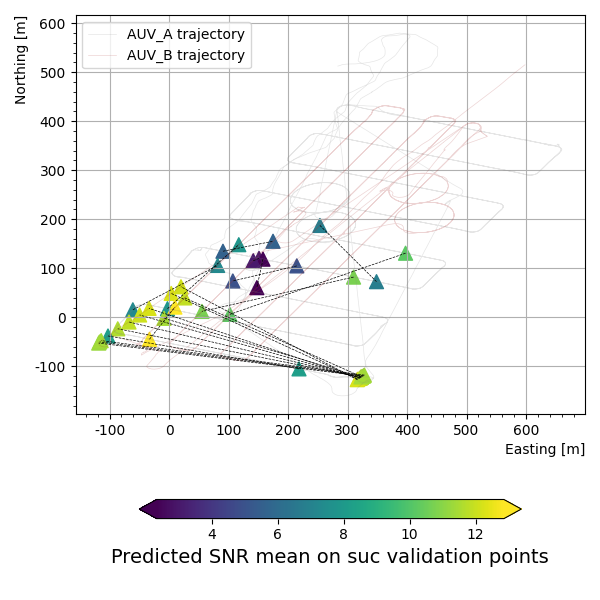}
  \end{subfigure}
  %\centering
  \begin{subfigure}[b]{0.23\textwidth}
  %\centering
  \includegraphics[width=\textwidth]{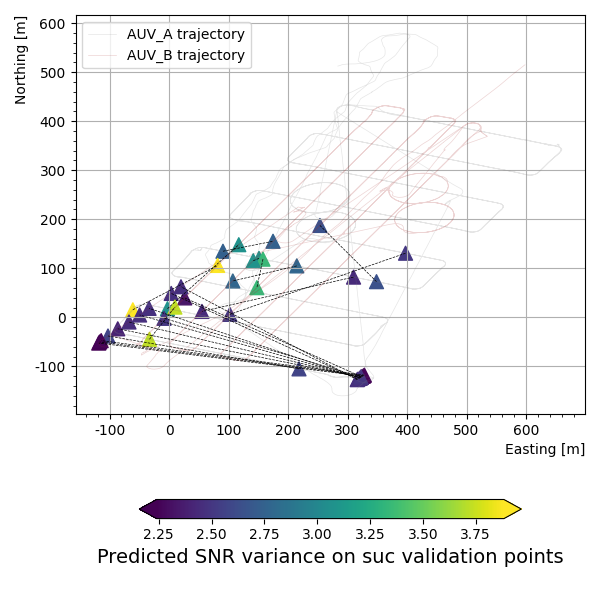}
  \end{subfigure}  

  \centering
    \begin{subfigure}[b]{0.23\textwidth}
  %\centering
  \includegraphics[width=\textwidth]{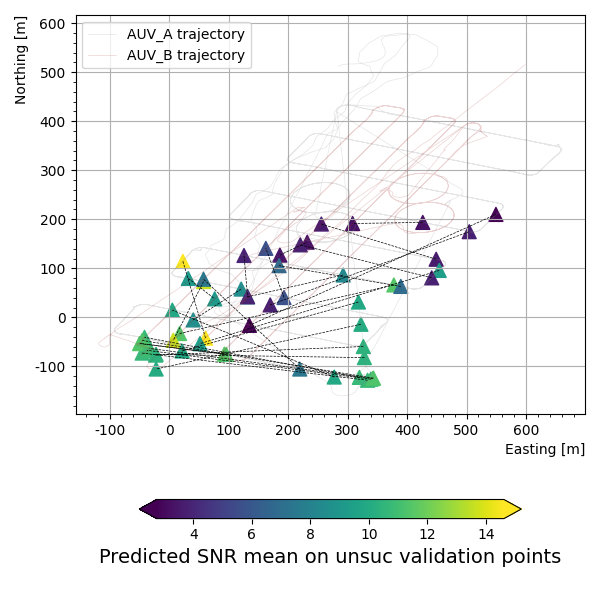}
  \end{subfigure}  
  %\centering
    \begin{subfigure}[b]{0.23\textwidth}
  % \centering
  \includegraphics[width=\textwidth]{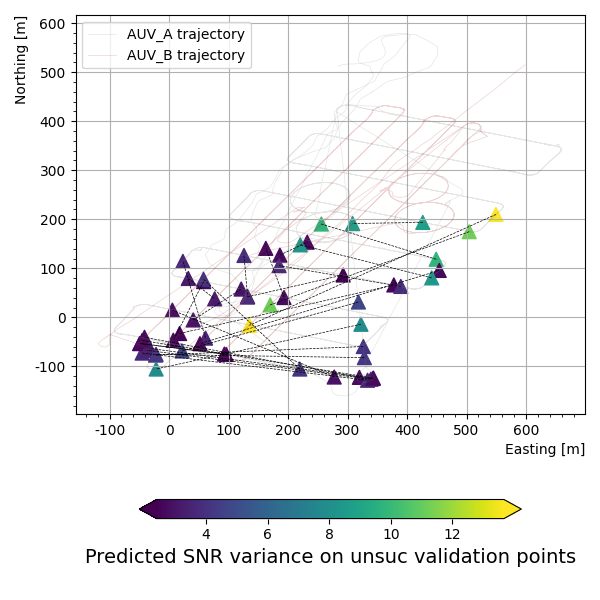}
  \end{subfigure}  
\caption{Predicted mean SNR(dB) and variance on successful (top row) and unsuccessful (bottom row) validation points using GPR}
\label{fig6}
\end{figure}

\begin{figure}[t!]
    \centering
  \begin{subfigure}[b]{0.23\textwidth}
  %\centering
  \includegraphics[width=\textwidth]{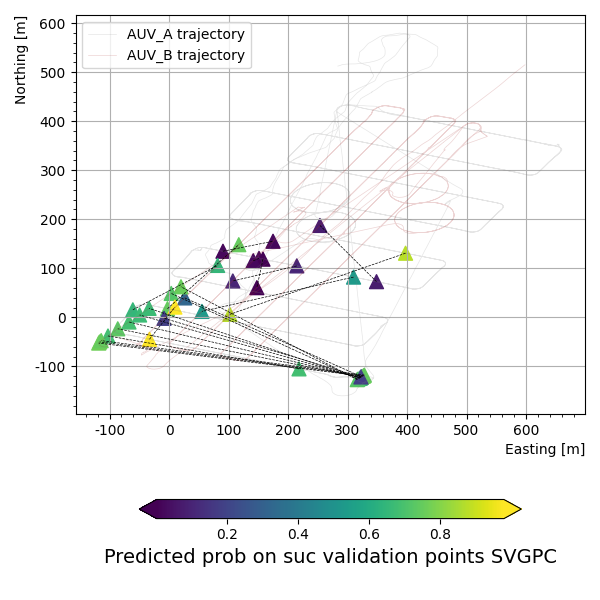}
  \end{subfigure}
  %\centering
  \begin{subfigure}[b]{0.23\textwidth}
  %\centering
  \includegraphics[width=\textwidth]{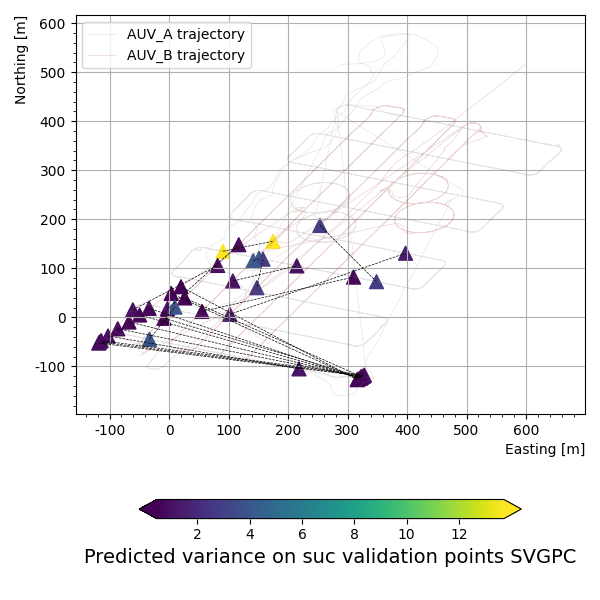}
  \end{subfigure}  

  \centering
    \begin{subfigure}[b]{0.23\textwidth}
  %\centering
  \includegraphics[width=\textwidth]{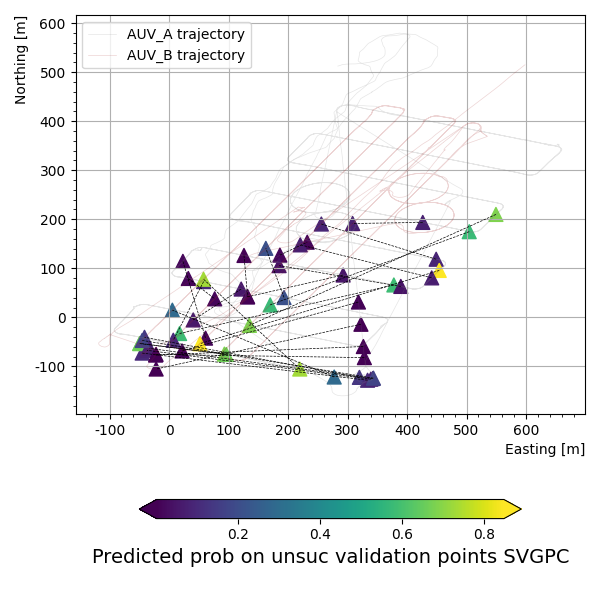}
  \end{subfigure}  
  %\centering
    \begin{subfigure}[b]{0.23\textwidth}
  % \centering
  \includegraphics[width=\textwidth]{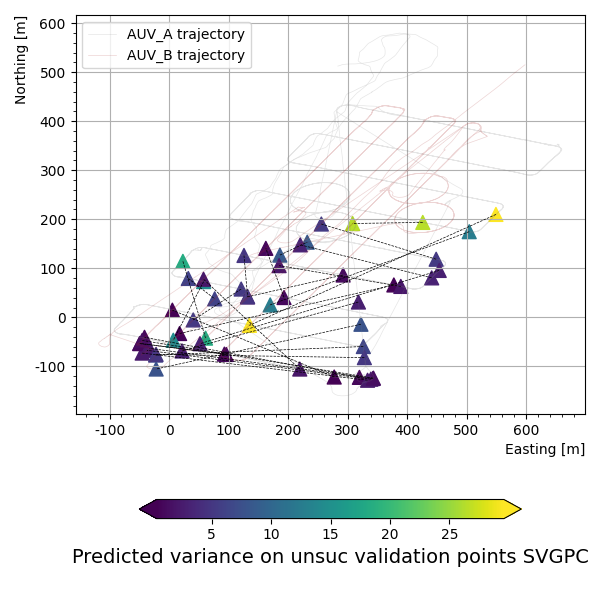}
  \end{subfigure}  
\caption{Predicted probability and latent variance on successful (top row) and unsuccessful (bottom row) validation points using SVGPC}
\label{fig7}
\end{figure}

\begin{figure}[h!]
\centering
  \begin{subfigure}[b]{0.45\textwidth}
  \centering
  \includegraphics[width=\textwidth]{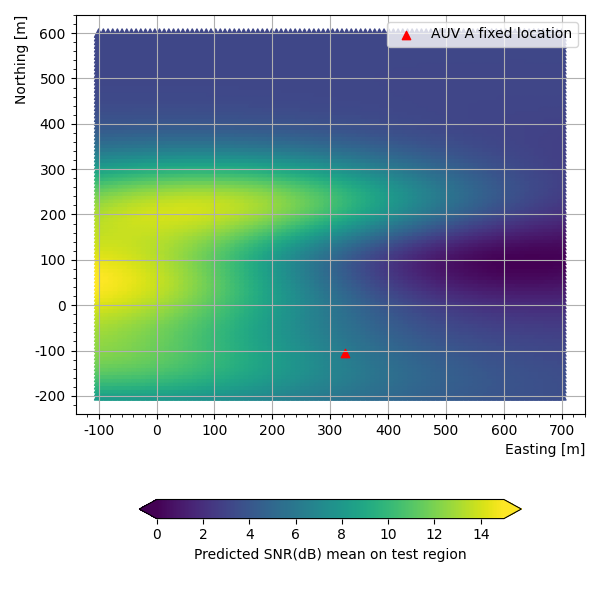}
  \end{subfigure}
  \begin{subfigure}[b]{0.45\textwidth}
  \centering
  \includegraphics[width=\textwidth]{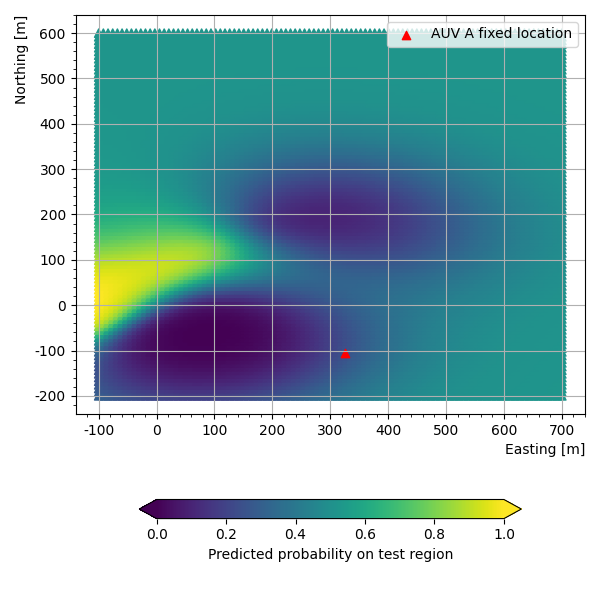}
  \end{subfigure}  
\caption{Prediction of mean and probability on fixed region using GPR and SVGPC}
\label{fig8}
\end{figure}

To illustrate the advantages of classification, we extract one validation set consisting of successful and unsuccessful validation points, as shown in Fig~\ref{fig5}. We present the performance for both GPR and SVGPC on this validation set after training, as indicated in Fig~\ref{fig6}, which include both mean and variance predictions for regression. For SVGPC in Fig~\ref{fig7}, we plot the corresponding probability and learned latent function variance. Clearly, we can see that in predicting successful communication events on the validation set, both GPR and SVGPC perform similarly, but GPR fails to predict unsuccessful communication events, especially around the region (340, -120) on the map, since it contains a significant number of both successful and unsuccessful communication events. Training with only successful events in GPR will predict this region as communication-safe, as shown in Fig~\ref{fig6}, where regression gives this region both high predicted signal-to-noise ratio (SNR) and low uncertainty, which is problematic. However, SVGPC in this validation set demonstrates good performance since it not only better predicts unsuccessful events (darker-colored dots) compared to GPR, but also estimates lower uncertainty on unsuccessful validation points in the region around (340, -120).  We did not compare the visualization of SVGPC with noise-input method since improvements are not very obvious on only one validation set, but improvements are shown in the results tables, which are averaged over 20 different validation sets. Finally, in Fig~\ref{fig8}, fix one AUV location(transmitting AUV A) on the map, indicated as red triangles on the figures, and generated heat maps on predicted SNR and probability for GPR and SVGPC. Here if we refer to SNR threshold $9.8dB$ and probability threshold $0.5$ on \textbf{DataSet2}, compared with the ground truth validation set, we can clearly see SVGPC outperforms regression since classification successfully identifies the boundary around the location (0.0, 0.0), which contains both successful and unsuccessful communication points on ground truth validation sets, but the regression method fails.

\begin{table}[h!] 
\caption{Mission 1(\textbf{DataSet1})}
\begin{center} \label{tab.Mission1}
\begin{tabular}{|c|c|c|c|c|c|} 
\hline
\textbf{Metric} & GPR & SVGPR & GPC & SVGPC & NI-NN\\
\hline

\hline\hline
\textbf{$ratio_{9.8dB/0.5}$} & 0.6670 & 0.6591 & \textbf{0.6954} & 0.6830 & \textbf{0.6909}\\
\hline
\textbf{$NLL_{0.5}$} & - & - & 0.7180 & 0.6961 & \textbf{0.6083}\\
\hline
\end{tabular}
\end{center}
\end{table}

\begin{table}[h!] 
\caption{Mission 2(\textbf{DataSet2})}
\begin{center} \label{tab.Mission2}
\begin{tabular}{|c|c|c|c|c|c|} 
\hline
\textbf{Metric} & GPR & SVGPR & GPC & SVGPC & NI-NN\\
\hline

\hline\hline
\textbf{$ratio_{9.8dB/0.5}$} & 0.5667 & 0.5490 & \textbf{0.6049} & 0.6627 & \textbf{0.7068}\\
\hline
\textbf{$NLL_{0.5}$} & - & - & 0.7503 & 0.9317 & \textbf{0.6205}\\
\hline
\textbf{$ratio_{10.2dB}$} & 0.5745 & 0.5520 & - & - & -\\
\hline
\end{tabular}
\end{center}
\end{table}

\begin{table}[h!] 
\caption{Mission 3(\textbf{DataSet3})}
\begin{center}\label{tab.Mission3}
\begin{tabular}{|c|c|c|c|c|c|} 
\hline
\textbf{Metric} & GPR & SVGPR & GPC & SVGPC & NI-NN\\
\hline

\hline\hline
\textbf{$ratio_{9.8dB/0.5}$} & 0.4602 & 0.4880 & \textbf{0.7530} & 0.6795 & \textbf{0.7253}\\
\hline
\textbf{$NLL_{0.5}$} & - & - & 0.5999 & 1.1328 & \textbf{0.7044}\\
\hline
\textbf{$ratio_{0.1dB}$} & 0.7500 & 0.7337 & - & - & -\\
\hline
\end{tabular}
\end{center}
\end{table}

% \begin{table}[h!]
% \caption{Virginia Tech 690 AUV}
% \label{table5}
% \begin{center}
% \begin{tabular}{|c|c|}
% \hline
% \textbf{PARAMATER} & \textbf{SPECIFICATION}\\
% \hline

% \hline\hline
% \multirow{1}{7em}{Length} & 87.7''\\
% \hline
% \multirow{1}{7em}{Displacement} & 94.7lbs\\
% \hline
% \multirow{1}{7em}{Diameter} & 6.9 inches\\
% \hline
% \multirow{2}{7em}{Endurance} & 22 hrs @ 3.2\\
% & knots\\
% \hline
% \multirow{1}{7em}{Depth} & 500 meters\\
% \hline
% \multirow{6}{7em}{Communications} & 900MHz RF\\
% & modem; Wi-Fi,\\
% & acoustic\\
% & communication,\\
% & satellite\\
% & communication\\
% \hline
% \multirow{2}{7em}{Navigation} & DVl $+$ IMU\\
% &acoustic ranging\\
% \hline
% \multirow{2}{7em}{Mission sensor} & Ping DSP 3D\\
% & side-scan sonar\\
% \hline
% \end{tabular}
% \label{tab:description}
% \end{center}
% \end{table}

% \textbf{Old ICRA:}
% We show the efficacy of our approach to predict acoustic communication performance using a real dataset acquired from Virginia Tech 690 AUVs in Claytor Lake, near Dublin, Virginia, USA. During data collection, each AUV is equipped with an inertial navigation system that estimates the AUVs location and the uncertainty in its location estimate, and an underwater acoustic modem that transmits and receives communication packets. Each AUV records the time and the location of received and transmitted communication events. Furthermore, each AUV transmits on a predetermined schedule so that the other vehicle can infer when an unsuccessful communication event has occurred.  

% To simplify the presentation of numerical examples, we assume that both AUVs are in the same horizontal plane and ignore vertical displacement. The dataset has 120 successful and 135 unsuccessful communication events between two AUVs. A subset of successful and unsuccessful communication events are shown in Fig~\ref{fig1}. The black solid line denotes the  trajectory of the first AUV, and the red solid line denotes the  trajectory of the second vehicle. The location of the receiving and transmitting AUVs for each communication event are indicated by a triangle that are linked by a solid line .      

\section{Concluding Remarks}
We demonstrate that acoustic communication events among underwater vehicles can be modeled as Gaussian process binary classification, and we illustrate the distinct advantages of Gaussian process classification-related methods over Gaussian process regression in predicting acoustic communication events. We provide a modified neural network method inspired by the NIMGPC framework and show that accounting for location uncertainty in the communication data set can further improve the communication ratio and negative log-likelihood compared to standard sparse Gaussian process classification. Finally, we visualize the prediction performance of both GPR and SVGPC methods on one validation set and show that SVGPC achieves better prediction performance with low uncertainty on unsuccessful communication events.

This work establishes a computational framework by which a team of AUVs can generate a map of communication performance using communication data acquired in the field. Herein we show how this framework is applied to historical data. The next steps in our research are to conduct experiments where maps are generated by AUVs in real-time, and to extend our framework to the decentralized case where the communication and computational burden imposed on each AUV for generating such a map is reduced.

% \textbf{Old ICRA:} We show acoustic communication events among underwater vehicles can be modelled as sparse variational Gaussian process binary classification. We provide a modified neural network and a first-order method inspired by the NIMGPC framework. In experiments we show that accounting for location uncertainty in the communication data set has the potential to increase both prediction accuracy and predictive distribution. Finally we demonstrate the training model approximately matches real communication data. 

% This work establishes a computational framework by which a team of AUVs can generate a map of communication performance using communication data acquired in the field.  Herein we show how this framework is applied to historical data.  The next steps in our research are to conduct experiments where maps are generated by AUVs in real-time, and to extend our framework to the decentralized case where the communication and computational burden imposed on each AUV for generating such a map is reduced.    

\bibliographystyle{ieeetr}
\bibliography{ref.bib}
\end{document}